\documentclass[runningheads]{llncs}

\usepackage[T1]{fontenc}
\usepackage{graphicx}
\usepackage{xcolor}

\usepackage{amsfonts}
\usepackage{comment}
\usepackage{amsmath,stix}
\usepackage{subfigure}

\begin{document}
\title{Grasp Force Optimization as a Bilinear Matrix Inequality Problem: A Deep Learning Approach}
\titlerunning{Grasp Force Optimization as a BMI Problem: A DL Approach}
\author{Hirakjyoti Basumatary\inst{1} \and Daksh Adhar\inst{1} \and Riddhiman Shaw\inst{2}  \and Shyamanta M. Hazarika\inst{1}} 

\institute{Biomimetic Robotics and Artificial Intelligence Lab (BRAIL), IIT Guwahati, India \and University of Edinburgh, United Kingdom.\\
\email{23hirak@gmail.com}}
\maketitle              
\vspace{-2em}
\begin{abstract}
Grasp force synthesis is a non-convex optimization problem involving constraints that are bilinear. Traditional approaches to this problem involve general-purpose gradient-based nonlinear optimization and semi-definite programming. With a view towards dealing with postural synergies and non-smooth but convex positive semidefinite constraints, we look beyond gradient-based optimization. The focus of this paper is to undertake a grasp analysis of biomimetic grasping in multi-fingered robotic hands as a bilinear matrix inequality (BMI) problem. Our analysis is to solve it using a deep learning approach to make the algorithm efficiently generate force closure grasps with optimal grasp quality on untrained/unseen objects. 


\end{abstract}
%
%
%

\section{Introduction}


Grasping and manipulation represent fundamental challenges in the field of robotics, and their effective execution is paramount for the successful deployment of robotic systems across diverse applications. Within this context, the optimization of grasp force emerges as a critical research area, aiming to enhance the efficiency and adaptability of robotic grasping strategies. Grasp force optimization (GFO) involves the precise modulation of forces exerted by robotic hands or grippers during object manipulation, aiming to achieve secure and stable grasps while minimizing energy consumption and potential damage to objects. This pursuit is particularly crucial in dynamic and unstructured environments, where robots must adapt their grasp forces in response to variations in object properties and environmental conditions. The problem of grasp force optimization comes from grasp kinematics, where optimal contact forces are calculated, which satisfies some equilibrium and friction constraints, ensuring stable and force-closure grasps. While including the above constraints, the GFO problem turns into a second-order cone programming (SOCP) problem \cite{chen2021multi}. 


This paper looks at grasp force optimization, exploring deep learning techniques rather than traditional optimization methods. The benefit of using deep learning techniques is better generalization capacities against unknown object properties and grasping conditions. The insights gained from such optimization efforts not only refine the grasp quality in industrial settings but also hold the potential to significantly impact domains such as warehouse automation, object recognition, and human-robot collaboration. Section 2 talks about the background and literature of matrix inequalities and methodologies to solve them. Section 3 discusses the deep learning methodology to solve convex optimization. Section 4 then discusses the formulation of grasp force optimization as a convex optimization problem. Section 5 discusses the numerical experimentation and the results obtained. Section 6 then elaborates on the validation based on grasp quality index of the obtained grasp. Finally, we conclude the paper in section 7.

\section{Background}

\subsection{Matrix Inequalities in Optimization}
\label{converting_BMI_to_LMI}

Linear Matrix Inequalities (LMIs) represent a crucial tool in control theory and optimization. An LMI is typically formulated as-
\vspace{-0.5em}
\begin{equation}
    F(x)=F_0 + \sum\limits_{i = 1}^m {{x_i}{F_i} \preccurlyeq 0} 
    \label{LMI_1}
\end{equation}
where, $ x \in {\mathbb{R}^m} $, $F_i \in {\mathbb{R}^{n \times n}} $. The inequality essentially states that $F(x)$ is a positive definite matrix. Mathematically,
\begin{equation}
    z^TF(x)z > 0, \forall z \neq 0, z \in {\mathbb{R}^n}
    \label{LMI_2}
\end{equation}
Expanding on LMIs, Bilinear Matrix Inequalities (BMIs) encompass a more complex form, where the solution sought is a vector,

\vspace{-0.5em}

\begin{equation}
    x \in {\mathbb{R}^n}
    \label{BMI_1}
\end{equation}
\vspace{-0.5em}
subject to a bilinear inequality,
\begin{equation}
    {F_0} + \sum\limits_{i = 1}^N {{x_i}{F_i} + \sum\limits_{i = 1}^N {\sum\limits_{j = 1}^N {{x_i}{x_j}{F_{ij}} \succcurlyeq 0,} } } 
    \label{BMI_2}
\end{equation}
where, $F_0, F_i, F_{ij}$ are constant $m \times m$ symmetric matrices.
Semidefinite programming (SDP) serves as a method to resolve BMIs by transforming them into a rank-constrained LMI with an additional variable $X \in {\mathbb{R}^{N \times N}}$ \cite{ibaraki2001rank}. The goal then becomes to find

\begin{equation}
    x \in {\mathbb{R}^N},X \in {\mathbb{R}^{N \times N}}
    \label{BMI_3}
\end{equation}
subject to the constraints,

\begin{equation}
    {F_0} + \sum\limits_{i = 1}^N {{x_i}{F_i} + \sum\limits_{i = 1}^N {\sum\limits_{j = 1}^N {{X_{ij}}{F_{ij}} \succcurlyeq 0,} } } 
\end{equation}


\begin{equation}
    M: = \left[ {\begin{array}{*{20}{c}}
X&x\\
{{x^T}}&1
\end{array}} \right] \succcurlyeq 0,
\label{BMI_5}
\end{equation}

\vspace{-0.5em}

\begin{equation}
    rank(M)=1
    \label{BMI_6}
\end{equation}


Each occurrence of the bilinear terms in (\ref{BMI_1}) and (\ref{BMI_2}) has been replaced by $(i,j)$ of the elements of the matrix X. Without the rank constraint (\ref{BMI_5}), (\ref{BMI_3}) is an SDP of the form-

minimize:
\begin{equation}
\label{SDP_1}
    f(x) = {c^T}x
\end{equation}

subject to:
\vspace{-1em}

\begin{equation}
\label{SDP_2}
    A(x) = {A_0} + \sum\limits_{i = 1}^n {{x_i}{A_i} \preccurlyeq 0} 
\end{equation}
where $x \in {R^n},c \in {R^n},{A_i} \in {R^{m \times m}},i = 1,2,...,n$

One of the ways to solve the optimization problem with BMI constraints is to drop the rank constraint of (\ref{BMI_3}) and solve a sequence of SDP with LMI constraints (like in (\ref{LMI_1}))  as shown in \cite{ibaraki2001rank}. In this paper, we will discuss the optimization of this SDP program using a modern deep-learning neural network approach. 

\subsection{Traditional and Modern Approaches of Solving LMIs}


As discussed in \cite{dai2018synthesis}, BMIs pose a significant computational challenge, falling into the category of NP-hard problems. To address these, various solution approaches have been developed within the areas of SDP. These approaches include interior point methods, first-order methods, bundle methods, augmented lagrangian methods, etc. \cite{majumdar2020recent}. The problem of optimization with bilinear constraints, exemplified in the context of grasping, further introduces non-convexity into the equation. Solutions for such range from convex relaxation to global optimization methods like branch and bound/branch and cut, as well as heuristic and metaheuristic approaches.



Neural networks are used to solve SDPs because they can be implemented effectively during large-scale integration, solve optimization problems with time-varying parameters, dynamical techniques and numerical ODE techniques can be implemented effectively, and they have a fast convergence rate in real-time solutions, essential for online implementation. SDPs are difficult to solve using traditional techniques because the constraints cannot be handled efficiently. To determine whether a matrix is positive semidefinite, for example, we must compute its eigenvalues or all determinants of its primary submatrices. \cite{jiang1999recurrent} discusses about using recurrent neural networks (RNN) to solve SDP problem. RNN have been proposed to solve many optimization problems because it is seen that they perform effectively in real-time computation.  When a RNN of this kind is put into hardware, the dynamics' steady state is often attained in milliseconds. Hence it can be used for real-time computation. In this paper, we discuss the deep learning approach given in \cite{wu2023deep} and modify the methodology to apply in to grasp force analysis problems.  

\section{Methodology}


Optimization problem with linear inequality constraints is of the form: 

\begin{equation}
\begin{aligned}
\min_{x} \quad & \textbf{D}^{T}\textbf{x}\\
\textrm{s.t.} \quad & \textbf{Ax}-\textbf{b} \le 0\\
\end{aligned}
\end{equation}
The corresponding Lagrangian is calculated as follows
\begin{equation}
    \textbf{L(x,u)} = {\textbf{D}^T}\textbf{x} + {\textbf{u}^T}(\textbf{Ax} - \textbf{b})
\end{equation}
The corresponding Karush–Kuhn–Tucker (KKT) conditions, as given in \cite{wu2023deep}, are,
\begin{equation}
    \begin{aligned}
{\textbf{D} + {\textbf{A}^T}\textbf{u} = 0,}\\
{{\textbf{u}^T}(\textbf{Ax} - \textbf{b}) = 0,}\\
{\textbf{Ax} - \textbf{b} \le 0,}\\
{\textbf{u} \ge 0.}
\end{aligned}
\end{equation}

Considering the functions \( x(t): \mathbb{R} \to \mathbb{R}^n \) and \( u(t): \mathbb{R} \to \mathbb{R}^m \), which are dependent on the time variable \( t \), let \( y(t) \) be the concatenated vector given by:
\begin{equation}
 \textbf{y}(t) = \begin{pmatrix} \textbf{x}(t) \\ \textbf{u}(t) \end{pmatrix}  
\end{equation}
The system of ordinary differential equations (ODEs) that models the Linear Programming (LP) problem, as expressed in equation (11), adheres to the KKT conditions delineated in equation (13). This time-varying definition allows us to write y in terms of t and further define,

\begin{equation}
    \frac{{d\textbf{y}}}{{dt}} = \phi (\textbf{y}) = \left[ {\begin{array}{*{20}{c}}
{\frac{{d\textbf{x}}}{{dt}}}\\
{\frac{{d\textbf{u}}}{{dt}}}
\end{array}} \right] = \left[ {\begin{array}{*{20}{c}}
{ - \left( {\textbf{D} + {\textbf{A}^T}{{(\textbf{u} + \textbf{Ax} - \textbf{b})}^ + }} \right)}\\
{{{\left( {\textbf{u} + \textbf{Ax} - \textbf{b}} \right)}^ + } - \textbf{u}}
\end{array}} \right]
\label{ODE_KKT}
\end{equation}

Let the neural network model be defined as
\begin{equation}
    \hat y(t;w) = (1 - {e^{ - t}})\textbf{NN}(t;w),\,\,\,y({t_0}) = {y_0},\,\,\,t \in [0,T],
    \label{neural_network_model}
\end{equation}

The loss function then becomes
\begin{equation}
    E(w) = \frac{1}{{\left| T \right|}}\sum\limits_{{t_i} \in T} {l\left( {\frac{{\partial \hat y({t_i};w)}}{{\partial {t_i}}},\phi (\hat y({t_i};w))} \right)} 
\end{equation}

With the framework of ordinary differential equations (ODEs) in place, incorporating the LP structure and the KKT conditions, along with a defined neural network model and an associated loss function, we are well-positioned to solve the SDP. This training will be aimed at efficiently finding the optimal solution to the LP problem.

\section{Grasp Force Optimization Formulation}


\subsection{Force Closure Constraints}

Given a set of $n$ contact points $ \{ {x_i} \in {R^3},i = 1,...,n\} $ and their friction cones ${(c_i, \mu)}$, where
$c_i$ is the friction cone axis and $\mu$ is the friction coefficient, a
grasp is in force closure if there exists contact forces ${f_i}$ at
${x_i}$ within ${(c_i, \mu)}$ such that ${x_i}$ can resist arbitrary external wrenches. For a grasp to be force closure, it has to satisfy the following constraints:  


\vspace{-1em}

\begin{align}
\textrm{} \quad & \label{bilinear_1} GG' \succcurlyeq  \varepsilon  {I_{6 \times 6}}\\
  & \label{bilinear_2} Gf = 0,\\  
  \vspace{-1em}
  & \label{bilinear_3} {f_i}^T{c_i} > \frac{1}{{\sqrt {{\mu ^2} + 1} }}\left| {{f_i}} \right|,\\
  &  \label{bilinear_4} {x_i} \in S
\end{align}
where S is the object surface and
\vspace{-1em}
\begin{equation}
    G = \left[ {\begin{array}{*{20}{c}}
{{I_{3 \times 3}}\,\,\,{I_{3 \times 3}}\,\,...\,\,\,{I_{3 \times 3}}}\\
{{{[{x_1}]}_ \times }\,\,\,{{[{x_2}]}_ \times }\,...\,\,\,\,{{[{x_n}]}_x}}
\end{array}} \right],\,\,\,{[{x_i}]_x} = \left[ {\begin{array}{*{20}{c}}
0&{ - {x_i}^{(3)}}&{  {x_i}^{(2)}}\\
{{x_i}^{(3)}}&0&{ - {x_{(i)}}^{(1)}}\\
{ - {x_i}^{(2)}}&{  {x_i}^{(1)}}&0
\end{array}} \right] \in {R^{3 \times 3}}
\end{equation}

As we can see, (\ref{bilinear_2}) is bilinear. Solving the preceding BMIs using SDP as discussed in Section. \ref{converting_BMI_to_LMI} yields an optimal $x_i$ and $f_i$. 





\subsection{Solving the Grasp Analysis as LMIs}
\label{section_grasp_analysis}



Given a multi-fingered robot hand grasping an object with $k$ contact points,  a grasp map $G \in {R^{6 \times m}} $ corresponds, which transforms fingertip forces from the local contact frame to the resultant object wrenches.
\vspace{-0.5em}
\begin{equation}
    F = Gx
\end{equation}
where $x = {[{x_1}^T...{x_i}^T...{x_k}^T]^T} \in {R^m}$ is the contact wrench of the grasp, and each component specifies wrench intensities at $i^{th}$ contact. In order to maintain a stable grasp, the net contact force $F$ must balance the external load $g_o$ experienced by the object.
\vspace{-0.5em}
\begin{equation}
    F = Gx = -g_0
\end{equation}
Under the contact model, the friction cone constraint that defines the set of contact wrenches and friction law applicable at $i^{th}$ contact is given as:
\begin{equation}
    F{C_i} = \left\{ {{x_i} \in {R^{{m_i}}}|{x_{in}} \ge 0,{{\left\| {{x_{it}}} \right\|}_w} \le {x_{in}}} \right\}
    \label{friction_cone_constraints}
\end{equation}
where, $x_{in}$ is the normal component of the contact force at contact $i$, $m_i$ is 3 and ${\left\| {{x_{it}}} \right\|_w}$ denotes the weighted norm of the frictional components at contact $i$. For point contact with friction (PCWF) model (our case), the weighted norm is defined as:

\begin{equation}
    PCWF:{\left\| {{x_{it}}} \right\|_w}: = \frac{1}{{{\mu _i}}}\sqrt {{x_{i1}}^2 + {x_{i2}}^2} 
\end{equation}
The hand equilibrium states that:
\begin{equation}
    {J_h}^Tx - {\tau _h} =  - {\tau _{ext}}
\end{equation}
where, ${J_h}$ is the hand jacobian, $\tau_h$ is the actuator efforts, $\tau_{ext}$ is the external load. 

The joint effort constraints $\tau$ on the contact wrench vector are written as:
\begin{equation}
    \tau  = \left\{ {x \in {R^m}|{\tau ^L} \le {J_h}^Tx + {\tau _{ext}} \le {\tau ^U}} \right\}
    \label{joint_effort_constraints}
\end{equation}

The friction cone constraints given by (\ref{friction_cone_constraints}) can be written as a positive semidefinite constraint on a block diagonal matrix P:
\begin{equation}
    FC = \left\{ {x \in {R^m}|P(x) = Blockdiag\left( {{P_1},...,{P_i},...,{P_k}} \right) \succcurlyeq  0} \right\}
\end{equation}
where $P_i$ is defined as,
\begin{equation}
    PCWF:{P_i}: = \left[ {\begin{array}{*{20}{c}}
{{\mu _i}{x_{i3}}}&0&{{x_{i1}}}\\
0&{{\mu _i}{x_{i3}}}&{{x_{i2}}}\\
{{x_{i1}}}&{{x_{i2}}}&{{\mu _i}{x_{i3}}}
\end{array}} \right]
\end{equation}

As $P_i$ matrices are linear and symmetric, we can write these friction constraints as nonstrict LMIs:
\begin{equation}
    {P_i} = {S_{i0}} + \sum\limits_{j = 1}^{{m_i}} {{x_{ij}}{S_{ij}}} 
\end{equation}

\begin{equation}
    {P_i} = {S_{i0}} + {x_{i1}}{S_{i1}} + ... + {x_{i{m_i}}}{S_{i{m_i}}} \succcurlyeq 0
\end{equation}

\begin{equation}
    P(x) = \sum\limits_{l = 1}^m {{x_l}{S_l} \succcurlyeq 0} 
    \label{friction_constraints}
\end{equation}

\textbf{Define what is S}

The joint effort constraints can be partitioned into two linear inequality constraints:

\begin{equation}
    {J_h}^Tx + {\tau _{ext}} - {\tau ^L} \ge 0,\,\,\,\, - {J_h}^Tx - {\tau _{ext}} + {\tau ^U} \ge 0
\end{equation}
And the following LMIs can then be formulated as follows,
\begin{equation}
    {T^{L,U}}(x) = diag({\pm J_h}^Tx \pm {\tau _{ext}} \mp {\tau ^{L,U}})
\end{equation}
\begin{equation}
    {T^{L,U}}(x)  = {T_0}^{L,U} + \sum\limits_{l = 1}^m {{T_l}^{L,U}{x_l} \succcurlyeq 0} 
\end{equation}

 The above LMIs can be together written as a single constraint-
\begin{equation}
    T(x) = Blockdiag({T^L}(x),{T^U}(x)) = {T_0} + \sum\limits_{l = 1}^m {{T_l}{x_l} \succcurlyeq 0} 
    \label{joint_effort_constraints}
\end{equation}
Finally, the friction cone constraint (\ref{friction_constraints}) and the joint effort constraint (\ref{joint_effort_constraints}) can be written as:

\begin{equation}
    D(x) = Blockdiag (P(x),T(x)) = {D_0} + \sum\limits_{l = 1}^m {{D_l}{x_l} \succcurlyeq 0} 
\end{equation}


\section{Numerical Results}
\vspace{-1.5mm}
Having established the theoretical underpinnings of our approach by expressing the necessary conditions as matrix inequalities and defining the neural network model, we will now proceed to apply this methodology to solve numerical examples. In particular, we will solve two examples- one taken directly from \cite{wu2023deep}, and the other, using the PyBullet GraspIt \cite{pybullet-grasp} to solve the grasp analysis problem, 
\vspace{-1.5em}
\subsection{Benchmark Example Problem}
\begin{figure*}[ht!]
  \includegraphics[width= 0.65 \linewidth]{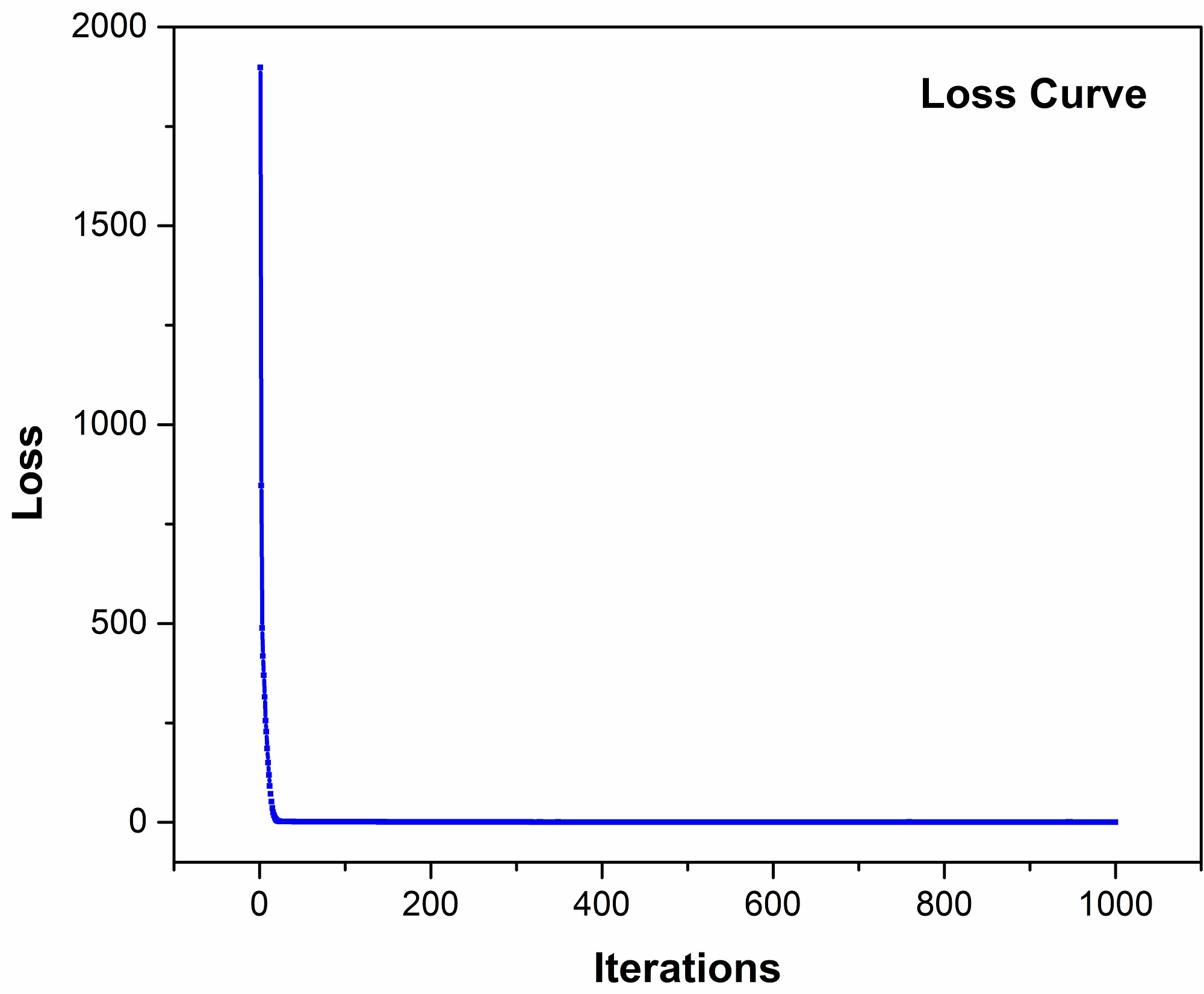}
  \centering
  \caption{The loss value in the first 1000 iterations}
  \label{example_1_loss_curve}
\end{figure*}

\vspace{-0.5em}
\begin{equation}
\begin{aligned}
\min_{x} \quad & -9.54x_1-8.16x_2-4.26x_3-11.43x_4\\
\textrm{s.t.} \quad & 3.18x_1+2.72x_2+1.42x_3+3.81x_4  \le 7.81 \\
\end{aligned}
\end{equation}
We make the ODE system according to (\ref{ODE_KKT}). By choosing the initial point to be all-zeroes and the time range to be [0,10], we construct a model based on (\ref{neural_network_model}) 
and train it in steps of 0.01. Our NN is layered as (1,100,5) which corresponds to 1 input neuron, 100 hidden layer neurons and 5 output neurons, with 'tanh' activation functions. As shown in Fig. \ref{example_1_loss_curve}, the loss decreases significantly and attains convergence. The final values received from our calculation are 
\(x_1= 0.999 , x_2= 0.579 , x_3= -0.228 ,x_4= 0.886.\)

\subsection{Grasp Analysis Problem}
To showcase our findings on an actual grasp problem, we use the PyBullet GraspIt \cite{pybullet-grasp} simulator for 3 fingered grasp from the Barrett hand as shown in Fig. \ref{grasp_planning}. During the simulation, a general grasp reveals the point of contact with the object (three), and we obtain the corresponding configuration of the hand for Jacobian calculation. This models our LMI problem as follows (discussed in Section. \ref{section_grasp_analysis}):
\begin{figure*}[h!]%
\centering
\subfigure[]{%
\label{object_grasp_pose}%
\includegraphics[height=4cm]{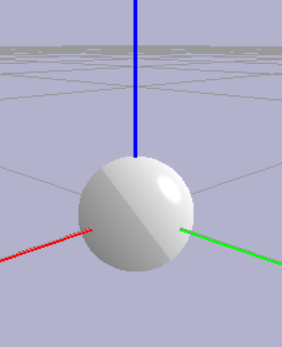}}%
\subfigure[]{%
\label{object_approaching}%
\includegraphics[height=4cm]{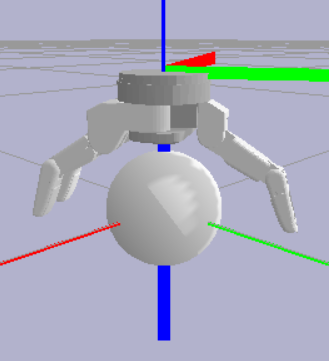}}%
\subfigure[]{%
\label{object_grasping_1}%
\includegraphics[height=4cm]{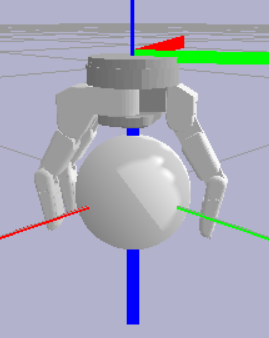}}%
\subfigure[]{%
\label{object_grasping_2}%
\includegraphics[height=4cm]{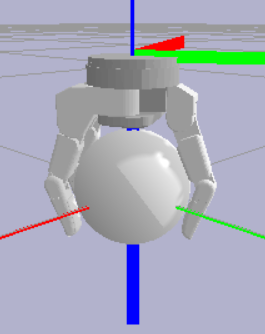}}%
\caption{Optimal Grasp Planning: (a) Object Position, (b) Initial Grasp Pose, (c) Intermediate Grasp Pose, (d) Final Grasp Pose }
\label{grasp_planning}
\end{figure*}

\begin{equation}
    \begin{aligned}
\min \quad & x_{11} + x_{12} + x_{13} + x_{21} + x_{22} + x_{23} + x_{31} + x_{32} + x_{33} \\
\textrm{s.t.} \quad & 1.7744x_{11} + 1.8984x_{12} + 1.5x_{13} + 2.1994x_{21} + 1.858x_{22} + \\ 
\quad & 1.5x_{23} +  1.8642x_{31} + 1.7924x_{32} + 1.5x_{33} + 600  \ge 0 \\
\end{aligned}
\end{equation}

\begin{figure*}[ht!]
  \includegraphics[width= 0.65 \linewidth]{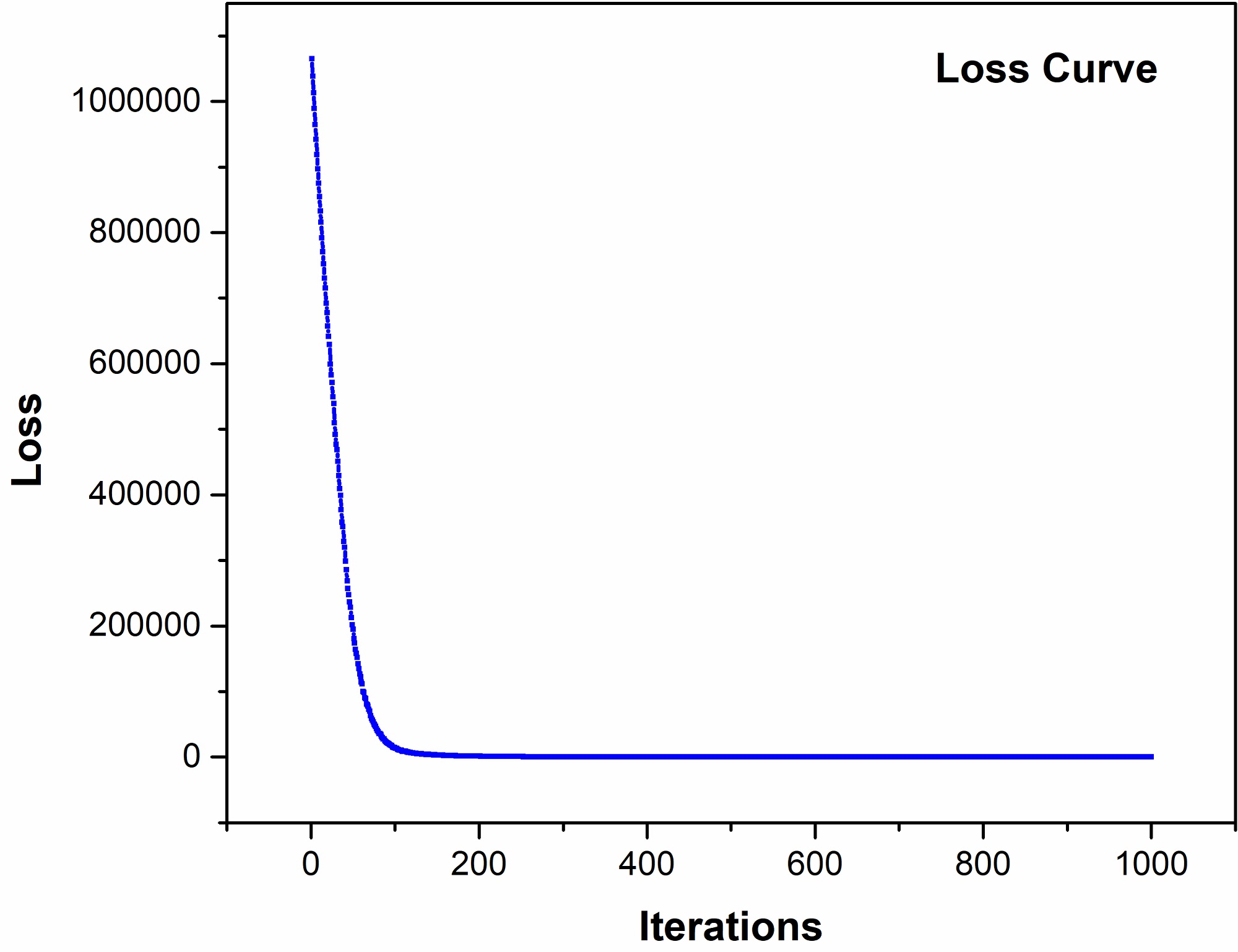}
  \centering
  \caption{The loss value in the first 1000 iterations}
  \label{example_2_loss_curve}
\end{figure*}

Solving using the same neural network structure, we obtain the solution set as follows 
 
$x_{11}= 36.601, x_{12}= -40.756, x_{13}= -38.333, x_{21}= -37.788, x_{22}= -35.586, x_{23}= -35.823, x_{31}= -39.289, x_{32}= -40.043, x_{33}= -34.805 $

where \(x_{ij}\) represents the \textbf{joint torque }values of the $j^{th}$ joints of the \(i^{th}\) finger. 
Once again, the loss function reduces with training, and the following curve is obtained as shown in Fig. \ref{example_2_loss_curve}. The optimal grasp planning is shown in Fig. \ref{grasp_planning}, where a Barrett hand is seen to grasp the object satisfying the friction cone constraints.

\section{Validation based on Grasp Quality Measure}

\begin{figure*}[ht!]
  \includegraphics[width= 0.65 \linewidth]{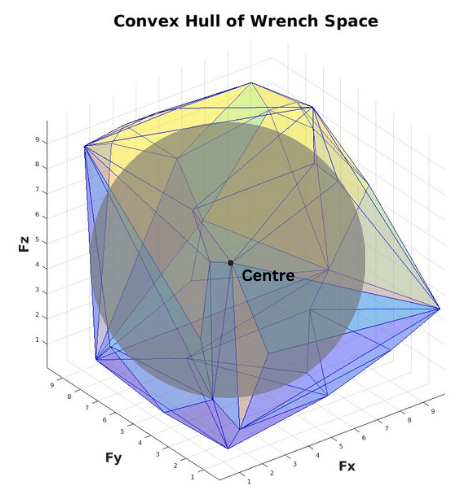}
  \centering
  \caption{Convex hull of the grasp wrench space of the optimal force input, showing the $q_{lrw}$}
  \label{convex_hull}
\end{figure*}

Finally, we analyse the grasp quality metric of the obtained grasp based on the optimal wrenches. We calculate the grasp quality $q_{lrw}$ i.e., the largest minimum resisted wrench, which is based on the convex envelope in the space of forces and wrenches that can be applied to the object during manipulation (shown in Fig. \ref{convex_hull}). The resultant wrench is given as:

\begin{equation}
    {\omega _0} = \sum\limits_{i = 1}^n {{\omega _i} = \sum\limits_{i = 1}^n {\sum\limits_{j = 1}^m {{\alpha _{ij}}\omega_{ij}} } } 
\end{equation}

where, ${\alpha _{ij}} \ge 0,\sum\limits_{i = 1}^n {\sum\limits_{j = 1}^m {{\alpha _{ij}} \le 1} } , \omega_i$ is wrench produced by finger $i$ at the $i^{th}$ contact and can be expressed as the poitive linear combination of the wrenches $\omega_{ij}$ produced by forces $f_{ij}$ ($m$ friction cone edges)

The convex hull of the wrenches is given as:

\begin{equation}
    P = CH\left( {\bigcup\limits_{i = 1}^n {\left\{ {{\omega _{i1}},...,{\omega _{im}}} \right\}} } \right)
\end{equation}

 The largest minimum resisted wrench grasp quality is given as:

\begin{equation}
    {q_{lrw}} = \mathop {\min }\limits_{\omega  \in \delta P} \left\| \omega  \right\|  
\end{equation}

where, $\delta P$ is the boundary of P

The $ {q_{lrw}}$ value in our case came out to be 6.7 (Fig. \ref{convex_hull}). This validates that our grasp is force closure, because it satisfies the criteria that the largest minimum resisted wrench needs to be positive to be force closure and to resist arbitrary torques and forces \cite{mnyussiwalla2022evaluation}. 

\section{Conclusion}

\vspace{-1em}


In conclusion, this paper has explored the paramount role of grasp force optimization in the realm of robotic manipulation, with a specific focus on leveraging deep learning approaches. The research presented herein underscores the significance of harnessing the capabilities of deep learning, to solve grasp force analysis with inequality constraints. The integration of neural networks has showcased promising results in optimizing grasp forces based on satisfying the force closure constraints. As the field continues to evolve, the application of deep learning in grasp force optimization elevates the performance of robotic hands to generalize across wider range of unseen grasping scenarios. The findings presented herein contribute to the ongoing research on the intersection of deep learning and robotic manipulation, paving the way for more sophisticated and context-aware grasping solutions in the ever-evolving landscape of robotics.

\vspace{-1em}
%
%
%
\bibliographystyle{splncs04}
\bibliography{mybibliography}
%

\end{document}